%% file: ijcai17.tex
\documentclass{article}
\usepackage{ijcai17}

\usepackage{times}

\author{
  Sibi Venkatesan, James K. Miller, Jeff Schneider \and Artur Dubrawski\\
  AutonLab, Robotics Institute, Carnegie Mellon University, Pittsburgh, PA\\
  \{sibiv, schneide, awd\}@cs.cmu.edu, mille856@andrew.cmu.edu}

\title{Scaling Active Search using Linear Similarity Functions}

\usepackage[small]{caption}
\usepackage{amssymb}
\usepackage{amsmath}
\usepackage{amsfonts}
\usepackage{graphicx}
\usepackage{bbold}
\usepackage{algorithmicx}
\usepackage[ruled]{algorithm}
\usepackage{algpseudocode}

\algnewcommand\algorithmicinput{\textbf{Input:}}
\algnewcommand\Input{\item[\algorithmicinput]}
\newtheorem{theorem}{Theorem}[section]

\newtheorem{lemma}[theorem]{Lemma}
\renewcommand{\L}{\mathcal{L}}
\newcommand{\U}{\mathcal{U}}
\newcommand{\K}{\mathcal{K}}

\newcommand{\inv}[1]{#1^{-1}}

\newcommand\blfootnote[1]{%
  \begingroup
  \renewcommand\thefootnote{}\footnote{#1}%
  \addtocounter{footnote}{-1}%
  \endgroup
}

\begin{document}

\maketitle

\renewcommand{\arraystretch}{1.5}

\blfootnote{Corresponding Author: Sibi Venkatesan (sibiv@cs.cmu.edu)}
\blfootnote{Published in Proceedings of IJCAI 2017.}

\input{abstract}
\input{introduction}
\input{relwork}
\input{problemstatement}
\input{approach}
\input{analysis}
\input{experiments}
\input{conclusion}

\bibliographystyle{named}
\bibliography{ijcai17}

\input{appendix}

\end{document}

%% file: abstract.tex
\begin{abstract}\label{sec:abstract}
Active Search has become an increasingly useful tool in information retrieval problems where the goal is to discover as many target elements as possible using only limited label queries.
With the advent of big data, there is a growing emphasis on the scalability of such techniques to handle very large and very complex datasets.

In this paper, we consider the problem of Active Search where we are given a similarity function between data points.
We look at an algorithm introduced by Wang et al. \cite{wangASgraphs} known as Active Search on Graphs and propose crucial modifications which allow it to scale significantly.
Their approach selects points by minimizing an energy function over the graph induced by the similarity function on the data.
Our modifications require the similarity function to be a dot-product between feature vectors of data points, equivalent to having a linear kernel for the adjacency matrix.
With this, we are able to scale tremendously: for $n$ data points, the original algorithm runs in $O(n^2)$ time per iteration while ours runs in only $O(nr + r^2)$ given $r$-dimensional features.

We also describe a simple alternate approach using a weighted-neighbor predictor which also scales well.
In our experiments, we show that our method is competitive with existing semi-supervised approaches. We also briefly discuss conditions under which our algorithm performs well.

\end{abstract}

%% file: introduction.tex
\section{Introduction}\label{sec:introduction}
	With rapid growth of the digital world, we are often faced with the task of quickly discovering and retrieving objects of interest from a large pool of data available to us. 
    The task of finding specific pieces of information might require more sophisticated solutions than just key-word searches. 
	Interactive approaches like Relevance Feedback can often be more effective, where an algorithm requests a user's feedback on its results in order to improve.
    Active Search is an example of such an approach: it discovers targets by asking the user for information it considers useful.
    With user feedback, Active Search algorithms iteratively build a model of what constitutes relevant information.
	This carries two potential benefits in information retrieval problems: 
    (1) these approaches need less labeled data and (2) they can focus on building a model of only the target class.
    The second point is useful for problems in which we are searching for the proverbial needle in a haystack.
    If there are relatively few targets, it is important to focus on modeling and identifying only those points.
	These approaches could be effective in real-world domains like product-recommendation and drug-discovery.

	In this paper, we look at the problem of Active Search given a similarity function between data points.
	This function induces a graph over our data, with edge-weights as the similarities between points.
	We consider an existing approach of Active Search on Graphs by Wang et al. \cite{wangASgraphs} and make key modifications which allow us to scale substantially.
	While the original approach looks at purely graphical data, we consider data lying in a multi-dimensional feature space.
    The similarity function is taken to be some kernel over features vectors.
    The only requirement is it is finite dimensional with an explicit kernel space representation.
   	In other words, the similarity function can explicitly be computed as the dot-product in this space.
    
    The original algorithm requires $O(n^3)$ pre-computation time, $O(n^2)$ time per iteration and $O(n^2)$ memory for $n$ data points.
    Ours only requires $O(nr^2+r^3)$ pre-computation time, $O(nr + r^2)$ time per iteration and $O(nr + r^2)$ memory for $r$-dimensional feature vectors.
    While the original approach is not viable for datasets larger than around 20,000 points, our algorithm comfortably handles millions.

	We also describe a simple approach using weighted neighbors which also scales to large datasets.
    This approach uses a Nadaraya-Watson-esque estimator to propagate labels, and runs in $O(nr)$ time for initialization and each iteration.
    
    The contribution of this paper is the following: 
    We present non-trivial modifications to an existing Active Search approach, scaling it multiple orders of magnitude.
    We describe a simple alternate which also scales well.
    We also touch upon when our algorithm will perform well.
    
   	This paper is structured as follows.
    We describe the existing literature in Section \ref{sec:relwork}.
    We formally state the problem of Active Search in Section \ref{sec:problemstatement}.
    In Section \ref{sec:approach}, we describe the existing approach followed by our modifications.
	In Section \ref{sec:analysis}, we discuss conditions for good performance.
    We describe our experiments and discuss results in Section \ref{sec:experiments}.
    We conclude in Section \ref{sec:conclusion}, and mention related challenges and next steps.

%% file: relwork.tex
\section{Related Work}\label{sec:relwork}
    Over the past few years, there has been significant research done in semi-supervised active learning.
    Most of this research is driven towards learning good classifiers given a limited labeled data, as opposed to recovering target points.
    
    Guillory et al. \cite{guillory2009label} propose methods for selecting labeled vertex sets on a graph in order to predict the labels of other points.
    Cesa-Bianchi et al. \cite{cesa2013active} explore an active version of this where they consider the optimal placement of queries on a graph to make minimal mistakes on the unlabeled points.
    
    Zhu et al. \cite{zhu2003semi} propose a method to perform semi-supervised learning on graphs.
    They formulate their problem in terms of a Gaussian random field on the graph, and efficiently compute the mean of the field which is characterized by a harmonic function.
    They extend this in \cite{zhu2003combining} to make it active:
    given the above graphical construction, they query points using a greedy selection scheme to minimize expected classification error.
    Zhu et al. \cite{zhu2005harmonic} describe a scalable method to perform inductive learning using harmonic mixtures, while preserving the benefits of graph-based semi-supervised learning.
    
    There has also been some work on optimization-based approaches for semi-supervised classification.
    Melacci et al. \cite{melacci2011primallapsvm} propose a method they call LapSVM, which builds an SVM classifier using the graphical structure of the data.
    Zhang et al. \cite{zhang2009prototype} describe the Prototype Vector Machine which solves a similar objective as above, by approximating it using ``prototype'' vectors which are representative points in the data.
    Liu et al. \cite{liu2010large} introduce an approach which also considers representative samples from the data called Anchors. 
    They construct an ``Anchor Graph'', and make predictions in the main graph based on weighted combinations of predictions on Anchors.
    
    Ma et al. \cite{maactive} describe new algorithms which are related to the multi-armed bandit problem to perform Active Search on graphs.
    Their algorithms are based on the $\Sigma$-optimality selection criterion, which queries the point that minimizes the sum of the elements in the predictive covariance as described in \cite{ma2013sigma}. 
    Kushnir \cite{kushnir2014active} also incorporate exploration vs. exploitation in their work on active transductive learning on graphs.
    They do this by considering random walks on a modified graph which combines the data distribution with their label hypothesis, allowing them to naturally switch from exploring to refinement.

    There have also been Active Search approaches which focus on recall instead of classification.
    Garnett et al. \cite{garnett2012bayesian} perform Active Search and Active Surveying using Bayesian Decision theory. 
    Active Surveying seeks to query points to predict the prevalence of a given class.

    Closely related to our work is that of Wang et al. \cite{wangASgraphs} where they perform Active Search on graphs.
	They select points by minimizing an energy function over the graph.
    They also emulate one-step look-ahead by a score reflecting the impact of labeling a point.
    Our work extends this with crucial modifications allowing us to scale to much larger data sets.

%% file: problemstatement.tex
\section{Problem Statement}\label{sec:problemstatement}

    We are given a finite set of $n$ points $X = \{x_1,\hdots,x_n\}$, and their unknown labels $Y=\{y_1,\hdots,y_n\}$ where $y_i \in \{0,1\}$. 
    We are also given a similarity function $\K(\cdot,\cdot)$ between points.
    We consider the case where this function is linear over some explicit feature transformation $\phi$:
    $\K(x_i,x_j) = \phi(x_i)^T \phi(x_j)$.
    This is analogous to the explicit kernel-space representation of some finite-dimensional kernel.
    This induces a graph over the data: the edge weight between $x_i$ and $x_j$ is given by $\K(x_i,x_j)$.
    
    Initially, we are given a small set of labeled points $\L_0$, while the remaining points are in the unlabeled set, $\U$. 
    Every iteration, we query one point in $\U$ for its label and move it to the labeled set $\L$. 
 	The goal is to find as many positive points as possible after $T$ iterations, where $T$ is a fixed budget for labeling points.

%% file: approach.tex
  \section{Approach}\label{sec:approach}
    \subsection{Background: Active Search on Graphs [ASG]}
      We briefly describe the algorithm introduced by Wang et al. \cite{wangASgraphs}.
      They interpret the data as a graph where the edge-weights between points is given by the similarity $\K$.
      Their method then uses a harmonic function $f$ to estimate the label of data points, inspired by the work done by Zhu et al. \cite{zhu2003semi}.
      This is done by minimizing the energy:
      \begin{equation}
        E(f) = \frac{1}{2}\sum\limits_{i,j} \K(x_i,x_j)(f(x_i) - f(x_j))^2
      \end{equation}
      The function $f$ serves as the primary measure for querying a point to label.
	  If $\K(\cdot,\cdot)$ is positive and semi-definite, the optimal solution $f^*$ can be interpreted intuitively through random walks on the graph: 
      $f^*(x_i)$ is the probability that a random walk starting at the point $x_i$ reaches a positively labeled point before a negatively labeled point.
      The following is a brief explanation:
      Here, for simplicity, we take $f^*$ to be the vector where $f^*_i := f^*(x_i)$.
      Setting the gradient of the energy to 0, we get at optimum $f^* = \inv{D} A f^*$ where $A$ and $D$ are the adjacency and diagonal degree matrices respectively.
      The rows of $\inv{D} A$ are exactly the transition probabilities from each node on the graph, and thus, the entire matrix can be interpreted as the transition matrix of the random walk.
      The interpretation of $f^*$ directly follows from this.

      Wang et al. then describe a problem they call hub-blocking, where a \textbf{negatively} labeled point is the center of a hub structure connected to multiple \textbf{positive} but unlabeled points.
	  Discovering another positive elsewhere in the graph will not help discover the positive unlabeled nodes in the hub, as they are blocked off by the negatively labeled hub center.
      To overcome this, they propose a soft-label model:
      every labeled point is now connected to a pseudo-node which holds the label instead.
      A random walk now terminates only when reaching the pseudo-node of a labeled point.
      A similar augmentation incorporates prior probabilities:
      a pseudo-node is attached to every unlabeled point and holds the prior probability of being positive.
      The transition probability from a point to its psuedo-node is constant across labeled or unlabeled points.

	  The following is the resulting energy function over $f$: 
      \begin{multline}
        E(f) = \sum\limits_{i \in \L} (y_i-f_i)^2 D_{ii} + \\
        \lambda \left(w_0 \sum\limits_{i \in \U} (f_i-\pi)^2 D_{ii} + \sum\limits_{i,j} (f_i-f_j)^2 A_{ij}\right)
      \end{multline}
      where $A_{ij} = \K(x_i,x_j)$, 
      $D_{ii} = \sum_j \K(x_i,x_j)$, 
      and the regularizing constants $\lambda$ and $w_0$ depend on transition probabilities into pseudo-nodes.
      Explicitly, if $\eta$ and $\nu$ are transition probabilities into labeled and unlabeled pseudo-nodes respectively, then $\lambda = \frac{1-\eta}{\eta}$ and $w_0 = \nu$.
      The minimizer of the energy function can be solved by setting the gradient to 0. We assume, without loss of generality, that the labeled and unlabeled indices are grouped together. The minimizer is\footnote{Derived in the Appendix.}: 
      \begin{equation}
        f^* = \inv{(I - B\inv D A)}(I-B)y',
      \end{equation}
      $$B = \left[\begin{matrix} \frac{\lambda}{1+\lambda}I_\L & 0 \\ 0 & \frac{1}{1+w_0}I_\U \end{matrix}\right], \hspace{3mm} y' = \left[\begin{matrix} y_\L \\ \pi \end{matrix} \right]$$
      This solution can also be obtained by performing label propagation in the augmented graph.
      For simplicity of notation, $f^*$ will simply be denoted by $f$ moving forward.

	  To pick points for label queries, ASG uses a heuristic called the Impact Factor which looks at the change of $f$ values if a given unlabeled point was labeled as positive.
      $$IM_i = f_i \sum\limits_{j \in \{U \backslash i\}} (f^+_j - f_j)$$
	  The final selection criterion is $\arg\max_i f_i + \alpha IM_i$.
      With this, ASG iteratively queries labels and updates $f$ and $\mathit{IM}$.
	  ASG has an $O(n^3)$ time initialization and $O(n^2)$ time per-iteration.
      
    \subsection{Linearized Active Search [LAS]}\label{sec:linearAS}
      Here, we describe our algorithm.
      We now require feature vectors for our points.
      The similarity function is then assumed to be be linear in these features (or some explicit transformation of them).
      This requirement is often not too restrictive; in fact, some popular kernels can be approximated using a linear embedding into some feature space.
      For example, the RBF kernel can be approximated by Random Fourier Features \cite{rahimi2007random}.
      For simplicity, let $x_i$ itself represent the feature vector.
      The similarity between two points is then $\K(x_i,x_j) = x_i^T x_j$.

	  Here is the algorithm at a glance.
      At a high level, LAS is the same as ASG:
      \begin{itemize}
      \item \textbf{Initialization:}
        Initialize with starting label set $L_0$.
        Pre-compute relevant quantities which can be updated.
      \item \textbf{In each iteration:}
      	Request the next label with a selection criterion based on $f$ and $\mathit{IM}$.
		Update all relevant quantities given this label.
      \end{itemize}

      \textbf{Note:}
      As mentioned before, ASG requires purely graphical data as input, i.e. the graph adjacency matrix.
      LAS works with a different class of data, which lives in some multi-dimensional feature space.
      A graph is induced over the data by the similarity function.
      If the input to ASG and LAS is the same, the results will be identical.
      By ``the same'', we mean the adjacency matrix for ASG is the same as the one of the induced graph for LAS.
      In this case, $f, IM$ and the point queried will be identical at every iteration.

	  \alglanguage{pseudocode}
      \begin{algorithm}
       \caption{LAS: Linearized Active Search}
       \label{alg:LAS}
       \small
       \begin{algorithmic}
         \Input $X, \mathcal{L}_0, w_0, \lambda, \pi, \alpha, T$
         \State $\U \leftarrow \{x_1,\hdots, x_n\} \backslash \mathcal{L}_0$
         \State Initialize $K^{-1}$, $f$, $\mathit{IM}$
         \For {$i = 1 \to T$}
           \State Query: $x_{i} \leftarrow argmax_\U (f + \alpha IM)$
           \State Update $K^{-1}$, $f$, $\mathit{IM}$ with $x_{i}, y_{i}$
           \State Remove $x_{i}$ from $\U$
         \EndFor
       \end{algorithmic}
      \end{algorithm}        
     
     The pseudo-code is given in Algorithm \ref{alg:LAS}. We now discuss how a linear similarity function helps us update $f$ efficiently. 
	 \subsubsection{Initialization}

       The adjacency matrix is $A = X^T X$ where $X = [x_1 \hdots x_n]$, with $n$ points and $r$ features.
       Then, $D = diag (X^T X \mathbb{1})$.
       This gives us:
	   $$f = \inv{(I - R X^TX)}q$$
       $$R = B \inv{D}, \hspace{5mm} q = (I-B)y'$$
       Using the matrix inversion lemma, we get\footnote{Derived in the Appendix.}:
       \begin{equation}\label{eq:f_init}
         f = q + RX^T\inv{K}Xq
       \end{equation}
       \begin{equation}\label{eq:k_init}
         K = I-XRX^T
       \end{equation}
       This converts an $O(n^3)$ time matrix inverse in ASG into the $O(r^3)$ time inverse of $K$.
       For large datasets, we can expect $r \ll n$.
       Below, we show that we only need to invert $K$ once; its inverse can be efficiently updated every iteration.

       The initialization runs in $O(nr^2 + r^3)$ time for computing $\inv{K}$ and $O(nr^2)$ for computing $f$.
       Next, we describe our efficient updates to $K$ and $f$ given a new label.

     \subsubsection{Updates to $f$ on receiving a new label}

       We have $\inv{K} = \inv{(I-XRX^T)}$ at the previous iteration.
       Only one element in $R$ changes each iteration.
       Take superscript $^+$ to mean the updated value of a variable.
       We have:
       $$R^+ = R -\gamma e_i e_i^T$$
       where
       $\gamma = -\left(\frac{\lambda}{1+\lambda}-\frac{1}{1+w0}\right)\inv{D_{ii}}$
       and $e_i$ is the $i^{th}$ standard basis vector.
       Using the matrix inversion lemma:
       \begin{equation}\label{eq:k_update}
         \inv{(K^+)} = \inv{K} - \dfrac{\gamma(\inv{K}x_i) (\inv{K}x_i)^T}{1+\gamma x_i^T \inv{K} x_i}
       \end{equation}
       Only one element in $q$ changes: 
       $q^+_i = y_i \dfrac{1}{1+\lambda}$.
       Thus, the update to $f$ can be calculated as\footnote{Updates derived in the Appendix.}:
       $$f^+ = q^+ + R^+X^T\inv{(K^+)}Xq^+$$
       This takes $O(r^2 + rn)$ time per-iteration as it just involves cascading matrix-vector multiplications.

     \subsubsection{Impact Factor}\label{sec:imfactor}
       LAS also includes appropriate modifications for the initialization and updates of the Impact Factor which adhere to the improved running time.
        We do not describe these here as they are much more involved than those above, while not being fundamentally complicated.\footnote{Derived in the Appendix.}
        We also slightly changed the Impact Factor from ASG: we scaled $\mathit{IM}$ so that it has the same mean as the $f$ vector.
        This allows us to tune $\alpha$ without worrying about the magnitude of values in $\mathit{IM}$, which varies based on the dataset.
        
     \subsection{Weighted Neighbor Active Search [WNAS]}
     	Here, we briefly describe a simple and intuitive alternate approach for query selection which also scales well with large amounts of data.
        This approach is similar to the Nadaraya-Watson kernel regressor:
        $$f_i = \dfrac{\sum_{j \in \L} y_i \cdot \K(x_i,x_j)}{\sum_{j \in \L} |\K(x_i,x_j)|}$$
        
        The updates for $f$ for this approach are simple.
		We keep track of the numerator and denominator individually for each unlabeled point.
        Each time we get a new labeled point $x_i$, we can compute its similarity to all other unlabeled points efficiently as the following vector: 
        $$\K(X_\U, x_i) = X_\U^{\rm T} x_i$$
        We can then update the numerator and denominator of all unlabeled points directly from this vector.
        The numerators would be updated by adding $y_i\K(X_\U, x_i)$ and the denominators would be updated by adding $|\K(X_\U, x_i)|$.
        These computations require $O(nr)$ time for initialization and iteration.

%% file: analysis.tex
\section{Analysis of Active Search}\label{sec:analysis}

  \subsection{Good Similarity Functions for Active Search}\label{sec:goodsim}
	How do we know if our similarity function is ``good'' for our problem, i.e., under what conditions will it give us a high recall rate for a given dataset?
    Not all similarity functions are suited to a given problem, even if they provide non-trivial information.
    For example, consider a similarity function which, given two animals, outputs 1 if they share the same number of legs and 0 otherwise.
    This similarity function will be useful to distinguish human beings from cats but not cats from dogs.
    But the similarity function itself is not useless.

    Assume that the similarity function only takes non-negative values.
    This allows us to interpret them as unnormalized probabilities.
    $f$ can be written as\footnote{Derived in the Appendix.}:
    $$f = \inv{(D + P - A)}Py' \hspace{2mm} \text{ where } \hspace{2mm} P = \left[\begin{matrix}\dfrac{1}{\lambda} I_\L&0 \\ 0&w_0 I_\U\end{matrix}\right]D$$
	Let: $M = D + P - A$. Given $M\mathbb{1} = (D-A)\mathbb{1} + P\mathbb{1} = P\mathbb{1}$, we get:
    $$f - \pi \mathbb{1} = \inv{M}Py' - \pi\mathbb{1} = \inv{M}P(y' - \pi\mathbb{1})$$
    M has the same sparsity structure as A, as all its off-diagonal elements are the negative of those in A.
    Since $M$ is a diagonally dominant symmetric matrix with non-positive off-diagonal entries, it is a Stieltjes matrix.
    This means that its inverse is symmetric and non-negative.
    Grouping indices by their class without loss of generality, we have $\inv{M}$ as
    $$\inv{M} = \left[\begin{matrix}\widetilde{M}_{11}&\widetilde{M}_{12} \\\widetilde{M}_{21} &\widetilde{M}_{22} \end{matrix} \right]$$
    This gives us:
    \begin{equation}\label{eq:f_cond}
      f - \pi \mathbb{1} = v_P + v_N
    \end{equation}
    $$v_P = \left[ (1-\pi)\left[\begin{matrix}\widetilde{M}_{11}P_P\\0\end{matrix}\right]u_P - 
      \pi\left[\begin{matrix}\widetilde{M}_{12}P_N\\0\end{matrix}\right]u_N\right]$$
    $$v_N = \left[ (1-\pi)\left[\begin{matrix}0 \\ \widetilde{M}_{21}P_P\end{matrix}\right]u_P -
      \pi\left[\begin{matrix}0 \\ \widetilde{M}_{22}P_N\end{matrix}\right]u_N\right]$$
    where $u_P$ and $u_N$ are indicator vectors of whether the points are labeled or not, for the positive and negative points respectively. 
    We only need to look at labeled points since for any unlabeled point $x_i$, $(y_i'-\pi) = 0$.
    Here, $\pi$ can be interpreted as a \textbf{parameter} instead of the constant prior, measuring ``importance'' of labels:
    if $\pi$ is low, then we consider each received positive label as very informative and vice-versa.
      
	Equation \ref{eq:f_cond} says that if the elements in $\widetilde{M}_{12} = \widetilde{M}_{21}^T$ are small, then $f$ will better reflect the labels of points.
    But when are these off-diagonal elements small?
    We can show that if the cross-class similarities, or off-diagonal blocks of A, are low in a matrix-norm sense, then the same is true in $\inv{M}$:
	\begin{lemma}\label{lemma:blkdiag}
      Let $A = A_1 + A_2$ where $A_1$ is the block diagonal component of the similarity matrix and $A_2$ is the pure cross-similarity component.
        
      If $||A_2||_1 < \epsilon$, then $||\widetilde{M}_{12}||_1 < \left(\dfrac{1}{c\cdot d_{min}}\right)^2\epsilon$ 
      where $c=\min\{\frac{1}{\lambda}, w_0 \}$ and $d_{min}$ is the minimum degree in the graph.\footnote{Proved in the Appendix.}
    \end{lemma}
    Intuitively, a bound on $||\widetilde{M}_{12}||_1$ bounds the between-class similarity.
    Lemma \ref{lemma:blkdiag} then tells us that if our similarity respects the underlying label distribution, then the computed $f$ will do the same.
    This only gives us information within a given iteration of Active Search; it does not directly give bounds on errors when querying the highest node in $f$ every iteration.
    But it is a step towards understanding the relationship between the similarity function and the performance.

    \begin{figure*}[h]
      \centering
      \includegraphics[width=0.245\textwidth]{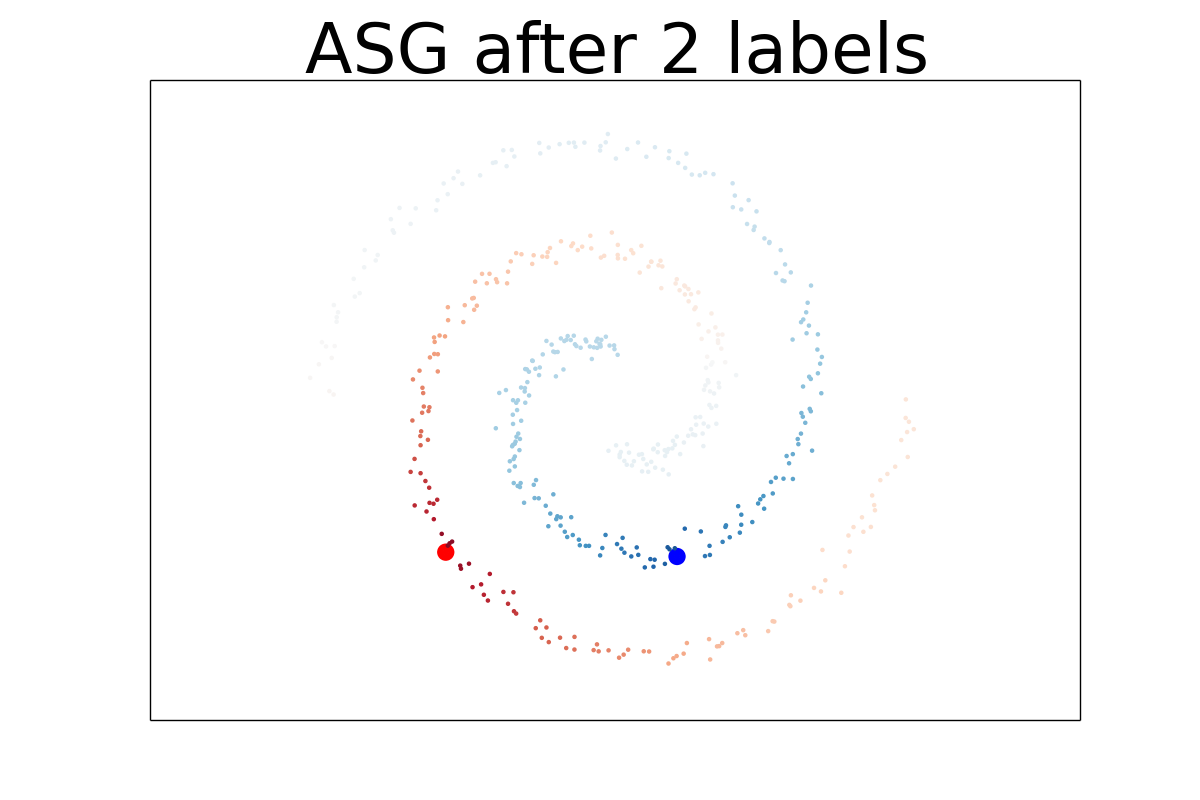}
      \includegraphics[width=0.245\textwidth]{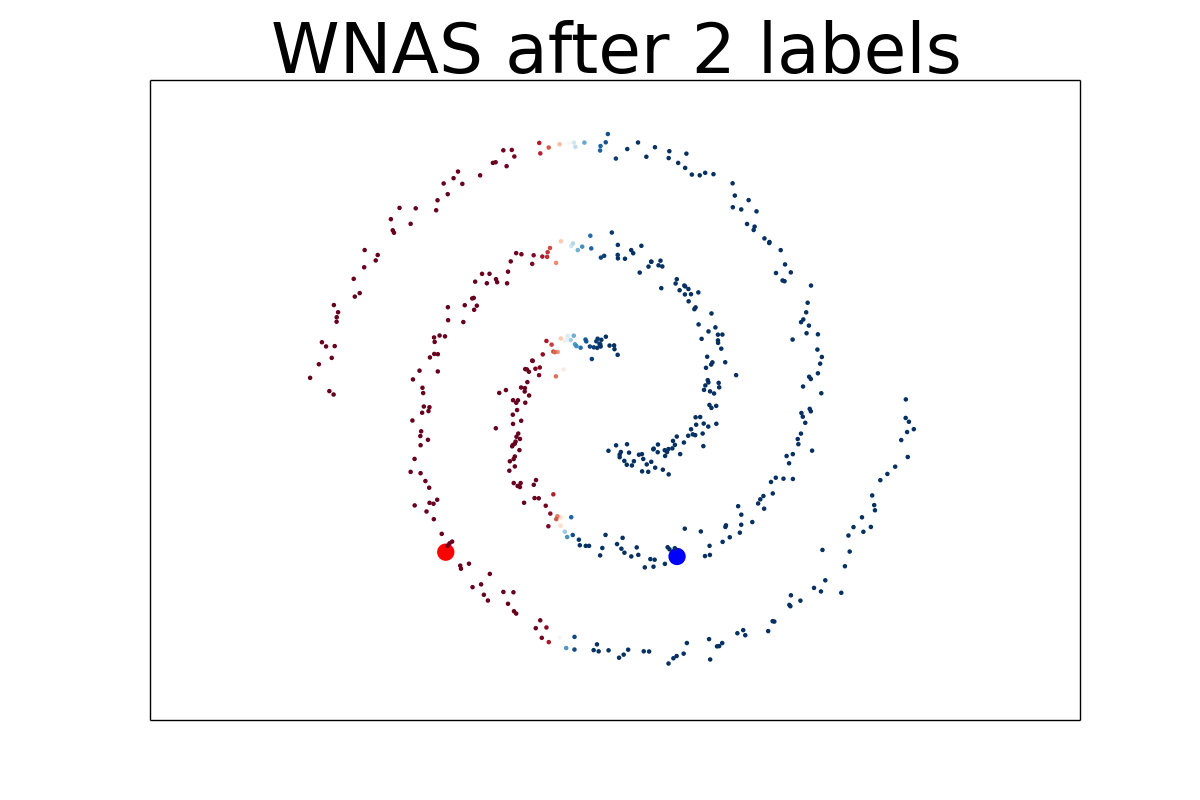}
      \includegraphics[width=0.245\textwidth]{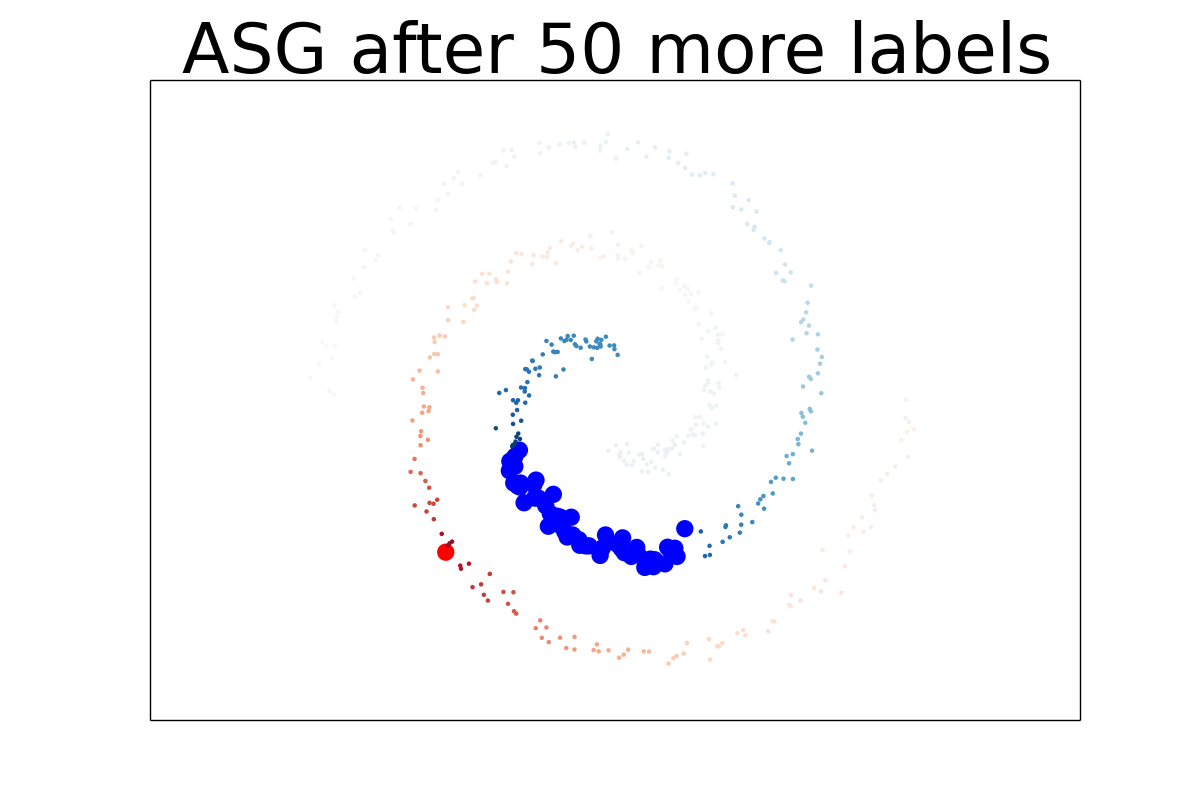}
      \includegraphics[width=0.245\textwidth]{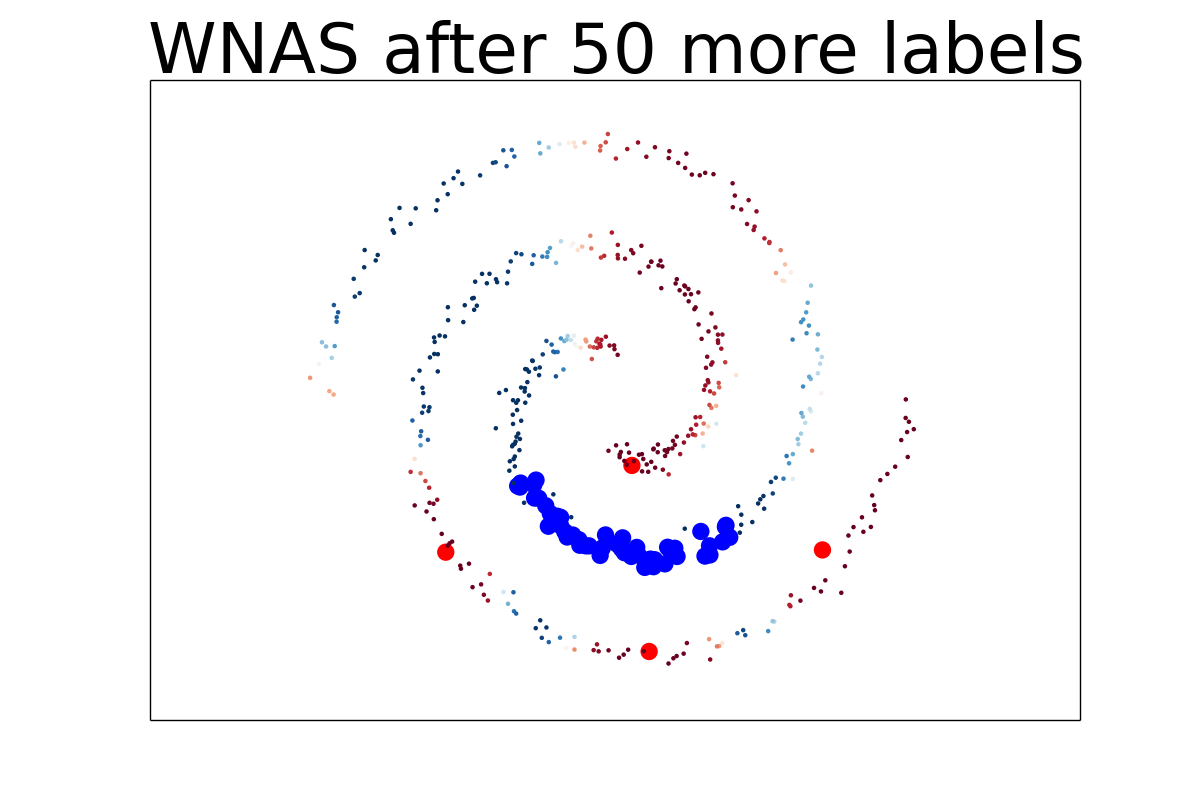}
 	\caption{These plots show the confidence of WNAS and ASG on unlabeled points after 2 and 50 labels. Blue corresponds to positive and red corresponds to negative, with intensity of color indicating the confidence. The big circles are labeled points.}
		\label{fig:ASGvsWNAS}
    \end{figure*}

  \subsection{Comparison of ASG/LAS and WNAS}\label{sec:lasvswnas}
    ASG (or equivalently LAS) and WNAS often have similar performance on recalling positive points.
    This is because, locally around the labeled points, both approaches propagate labels in a similar manner.
	The label confidences assigned by WNAS can also be interpreted as one step of a random-walk as follows.
    For each unlabeled point, consider the graph containing it, along with all labeled points.
    Its $f$ score is equivalent to the probability that a random walk starting from that point transitions into a positive in one step.
    
    However, while WNAS makes use of local structure of the graph around the labeled points, it does not effectively use the global connectivity structure of the graph.
	The relevance of this can be seen in the swiss-roll dataset in Figure \ref{fig:ASGvsWNAS}; where the inner blue roll is positive and the outer red roll is negative.
    The predictions of WNAS and ASG are similar around the labeled points, but very different away from them.
    The usefulness of WNAS' $f$ score diminishes rapidly moving away from the label set, unlike for ASG.
    
    Another note is that computing an equivalent Impact Factor for WNAS requires $O(n^2)$ computation, since we need to compute the similarity between every pair of unlabeled points.
    This makes the Impact Factor computation infeasible for large datasets, unlike LAS as discussed before.

%% file: experiments.tex
    \begin{figure*}
      \centering
      \includegraphics[width=0.33\textwidth]{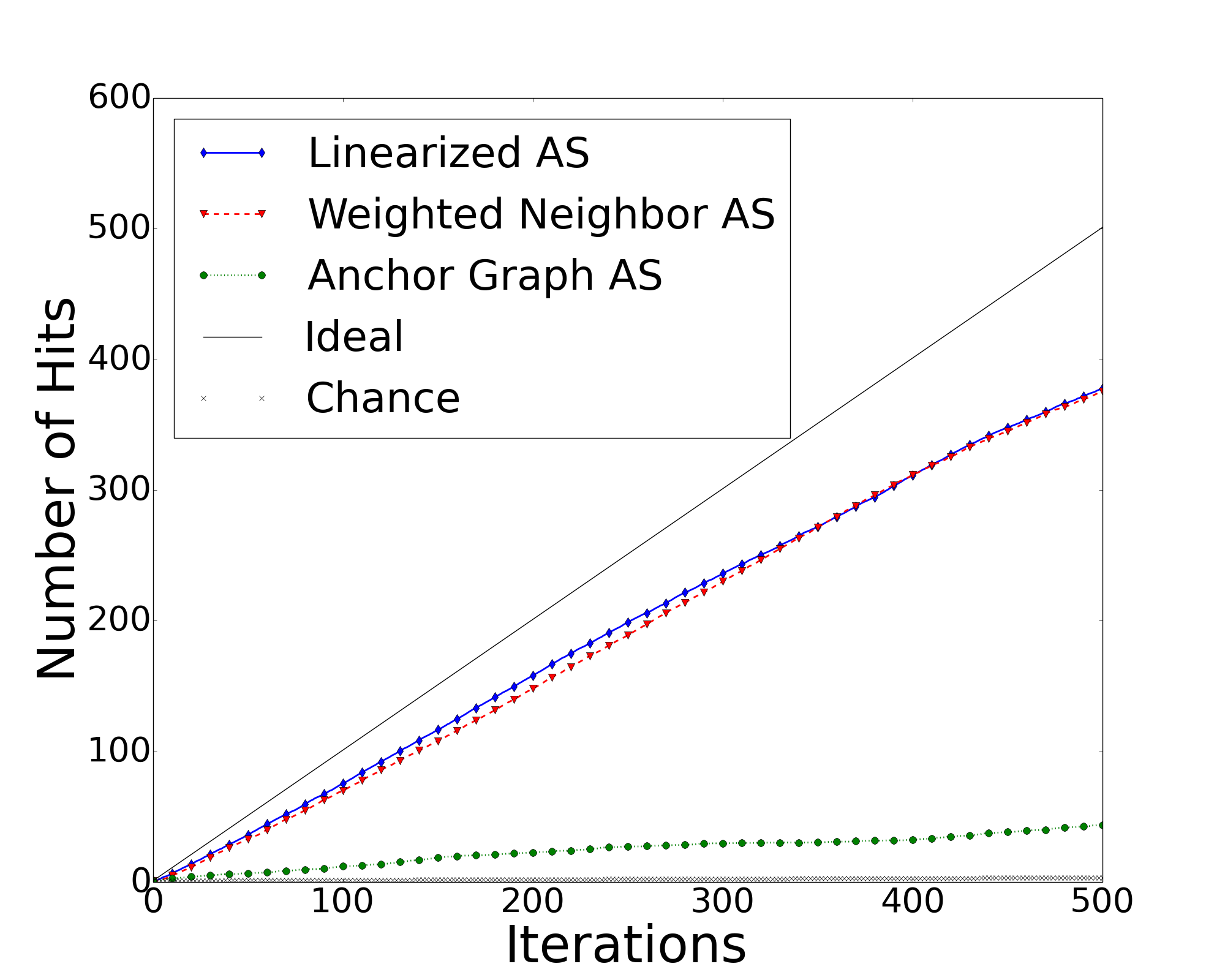}
      \includegraphics[width=0.33\textwidth]{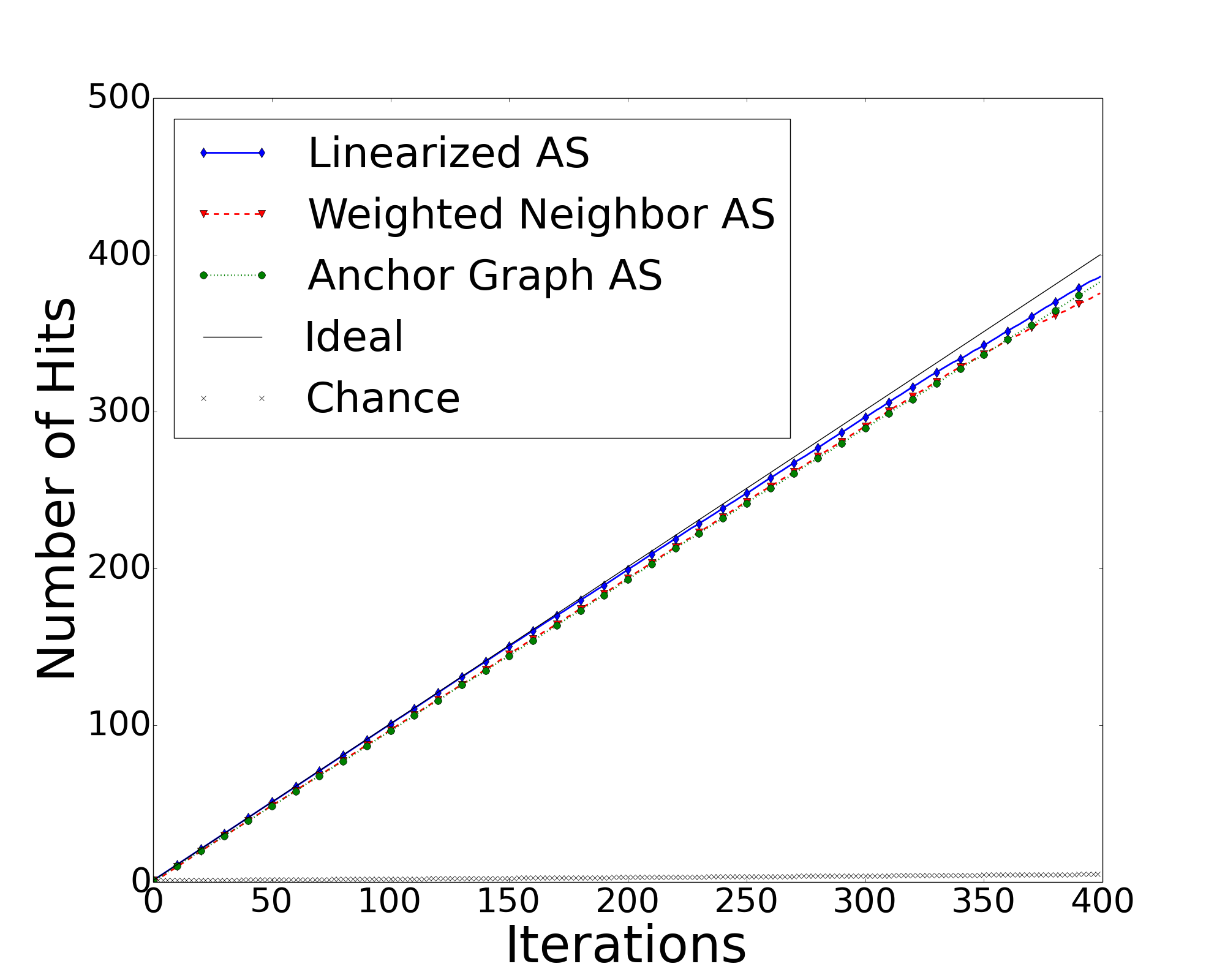}
      \includegraphics[width=0.33\textwidth]{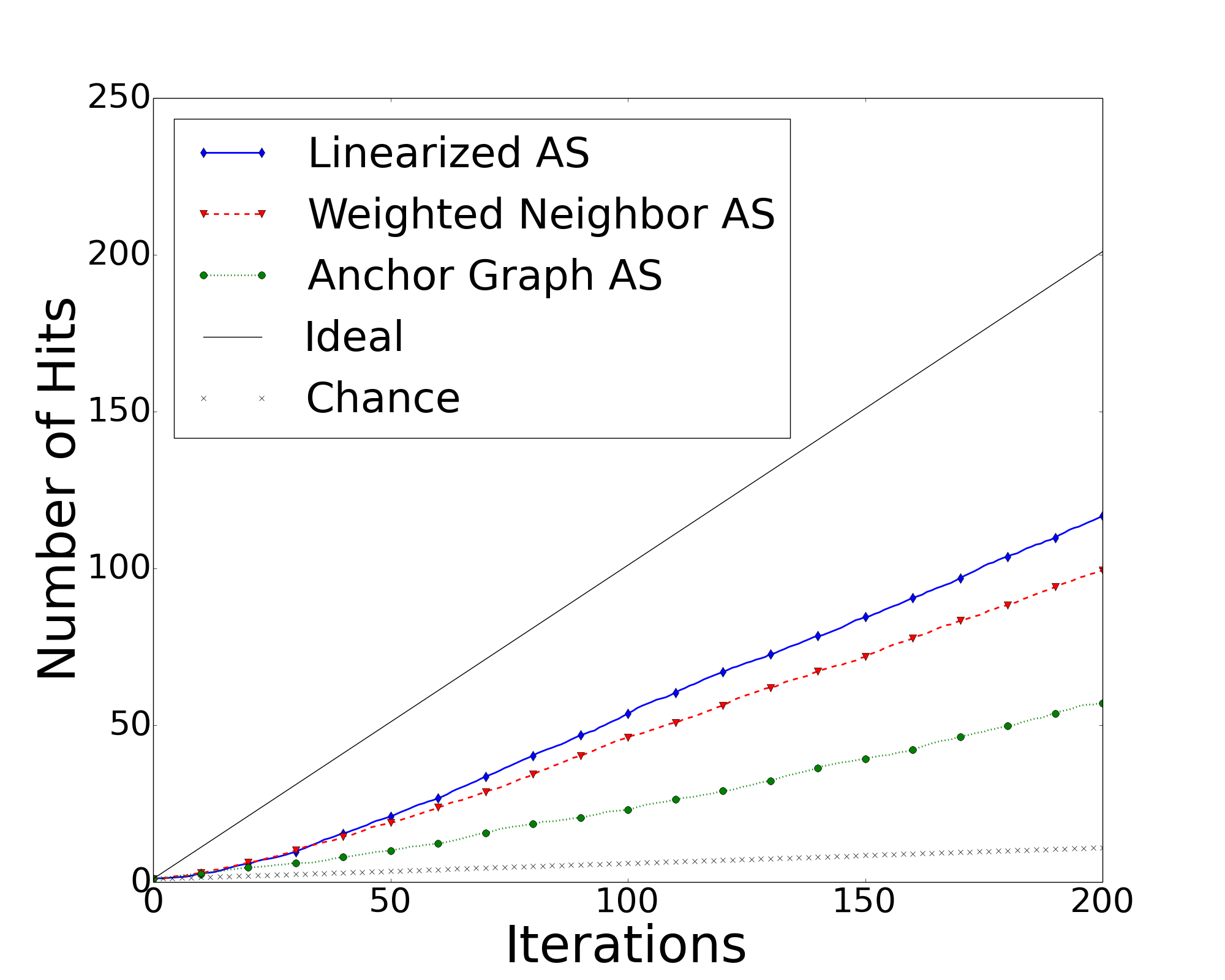}
	  \caption{These plots show recall vs. iteration averaged across 10 runs for LAS, WNAS and AGR, along with ideal and random recall. The left image is for CoverType, the middle image is for MNIST and the right image is for Adult.}
	  \label{fig:main_expts}
    \end{figure*}
    \begin{table*}[t]
      \centering
      \begin{tabular}{|c|cc|cc|cc|}
        \hline
        &\multicolumn{2}{c}{CoverType} & \multicolumn{2}{c}{MNIST} & \multicolumn{2}{c|}{Adult}\\
        \hline
        & 250 & 500 & 200 & 400 & 100 & 200 \\
        \hline
        LAS & \textbf{198.7 $\pm$ 32.0} & \textbf{377.8 $\pm$ 55.7} & \textbf{199.5 $\pm$ 1.0} & \textbf{386.4 $\pm$ 4.9} & \textbf{53.7 $\pm$ 11.7} & \textbf{116.7 $\pm$ 13.3} \\
        WNAS & 188.8 $\pm$ 21.5 & 375.7 $\pm$ 37.9 & 193.7 $\pm$ 3.0 & 379.8 $\pm$ 7.2 & 46.1 $\pm$ 16.3 & 99.4 $\pm$ 26.4 \\
        AGR & 27.2 $\pm$ 11.2 & 43.5 $\pm$ 11.8 & 192.8 $\pm$ 3.1 & 380.1 $\pm$ 4.0 & 23.1 $\pm$ 18.5 & 57.1 $\pm$ 39.2 \\
        \hline
      \end{tabular}
  	  \caption{This table shows mean recall $\pm$ standard deviation at the middle and last iteration for each algorithm and dataset.}
	  \label{table:main_expt}
    \end{table*}

\section{Experiments}\label{sec:experiments}
    We performed experiments on the following datasets:
	the CoverType and Adult datasets from the UCI Machine Learning Repository and MNIST.

	The Covertype dataset contains multi-class data for different forest cover types.
    There are around 581,000 points with 54-dimensional features.
    We take the class with the lowest prevalence of $0.47\%$ as positive.
    The data is unit normalized across features and a bias feature is appended to give 55 in total.
    Then, we project these onto a 550-dimensional space using Random Fourier Features \cite{rahimi2007random} to approximate an RBF Kernel.

	The Adult dataset consists of census data with the task of predicting whether a person makes over \$50k a year or not.
    It contains 14 features which are categorical or continuous.
    The continuous features are made categorical by discretization.
    Each feature is converted into a one-hot representation with $m$ binary features for $m$ categories.
    The features are then unit normalized.
    The positives are those making more than \$50k a year.
    We modified the dataset size to make the target prevalence 5\%.
    The final dataset has a~39,000 points.
    
    For the MNIST dataset, we combine the training, validation and testing sets into one.
    The 28x28 pixel images give us 784 features which are then unit normalized.
    We take the positive class to be the digit 1, and modified its prevalence to be 1\%.
    The final dataset has around 63,500 points.
    
  	We compare LAS and WNAS to \textbf{Anchor Graph Regularization} with Local Anchor Embedding [AGR] as described in \cite{liu2010large}\footnote{This was re-implemented in Python for our experiments.}.
    Their approach creates a proxy graph called the Anchor Graph which approximates the larger dataset; the labels given to points are then a weighted combination of the labels of the anchor points.
    Since this is a semi-supervised classification approach, we retrain it every iteration with all the data and known labels.
    We then use the confidence values for each unlabeled point to be positive as the $f$ value.
    This algorithm requires anchors to be computed beforehand.
    For this, we generated k-means over the transformed data points, with $k=500$ for each dataset.

	Our main experiment measured recall (number of positives found) over a fixed number of iterations for each dataset.
    For each dataset, 10 runs were performed starting with one randomly chosen positive as initialization.
    For LAS, we took $\alpha$ (the coefficient for the Impact Factor) to be the best from empirical evaluations.
    This was $10^{-6}$ for CoverType and Adult, and 0 for MNIST.
    $\pi$ was taken as the true positives prevalence.
	
    We also carried out smaller experiments over each dataset where we studied the predictive performance of LAS vs. WNAS immediately after initialization.
    Here, we randomly sampled 100 pairs of one positive and one negative point to initialize.
    Then, we reported the number of positives in the top 100 unlabeled points according to their $f$-values.
    These 100 pairs did \textbf{not} include ``bad'' initializations, where neither approach found any positives.

	\textbf{Note:} We did not compare our approach vs. purely graph based methods as in the \cite{wangASgraphs}
    Since our results are identical to ASG given the ``same'' data as described before, we only considered data with feature vectors.
 
\subsection{Results}
  Figure \ref{fig:main_expts} shows plots of the recall per iteration of LAS, WNAS and AGR for the different datasets.
  Table \ref{table:main_expt} shows mean recall and standard deviation of these experiments in the mid and final iteration.
  LAS and WNAS both have good performance in all three experiments.
  The CoverType dataset has high variance in estimates, likely because the data has many scattered positives which are not very informative during initialization.
  The algorithms would then take longer to discover the remaining positives.
  The MNIST data-set showed particularly good performance across the different approaches; all three approaches have near ideal recall.
  This is likely because the targets are tightly clustered together in the feature-space.
  The performance of AGR in the CoverType, though much better than random choice, is poorer than the other approaches.
  This is because AGR incurs significant overhead in the initialization of the algorithm.
  Computing k-means, followed by the weights and the reduced Laplacian of the Anchor Graph takes a few hours for CoverType.
  Furthermore, any change in the feature function used between the data points requires recomputation of the Anchor Graph.
  Due to this, we only used 500 Anchors even though it is a larger data-set.
  This poorer approximation of the data likely led to worse performance.

  Table \ref{table:one_positive} shows the comparison between LAS and WNAS given a single positive and negative point for initialization.
  As expected from our discussion in Section \ref{sec:lasvswnas}, LAS generalizes better with the unlabeled data.
  
  \textbf{Note:} We also conducted similar experiments on much larger datasets from the UCI Repository: the HIGGS dataset (5.5 million points) and the SUSY dataset (2.5 million points).
  We have not reported these results.
  These experiments were not any more informative than those above; they just served as a demonstration of scale.
  
  \begin{table}
    \centering
    \begin{tabular}{|c|c|c|}
      \hline
      Dataset (pos\%) & LAS & WNAS\\
      \hline
      Covertype (0.47\%) & \textbf{4.19} & 1.66\\
      MNIST (1.00\%) & \textbf{94.25} & 60.68\\
      Adult (5.00\%) & \textbf{27.25} & 17.29\\
      \hline
    \end{tabular}
    \caption{This table shows the average positives in the top 100 unlabeled points from the $f$-values of LAS and WNAS.}
    \label{table:one_positive}
  \end{table}

%% file: conclusion.tex
\section{Conclusion and Future Work}\label{sec:conclusion}

  In this paper, we proposed an algorithm to perform Active Search given a linear similarity function between data points.
  Through experiments, we demonstrate the scalability of our algorithm as compared to the original approach by \cite{wangASgraphs}, as well as good recall on different datasets.

  We also described an alternate, simple approach using a weighted neighbor estimator of labels.
  This approach also scales well to large datasets, but is not as generalizable given very little labeled information.
  It does perform comparably with our main approach in the recall problem.

  \subsection{Future Work}
  We require a ``good'' similarity function, or equivalently a good featurization, for our approach to perform well.
  A next step would be to learn a featurization simultaneously while performing Active Search.
  The challenge here is effective regularization with very little labeled data at the beginning.
  
  We have also not dealt with natural graphs in this paper, because of the restriction on our similarity function.
  But we know that every iteration of Active Search just uses label propagation to compute $f$.
  There exist methods, such as \cite{fujiwara2014efficient}, to perform efficient label propagation on large sparse graphs.
  Incorporating this into our approach along with appropriate Impact Factor computation would allow us to scale on natural graph datasets.

\section*{Acknowledgments}

Work partially supported by DARPA (FA8750-12-2-0324, FA8750-14-2-0244) and NSF (1320347).

%% file: appendix.tex
\appendix
    \section{Algorithm Derivations}
    \subsection{Derivation of minimizer of E($f$)}\label{sec:minEf}
      We can rewrite the energy function using matrices as given below. 
      The subscripts $\L$ and $\U$ represents portions of various quantities as belonging to the set of labeled and unlabeled points respectively. 
      Here, without loss of generality, we have rearranged the $f$-vector to group the labeled and unlabeled points.
      This re-arrangement will hold throughout this appendix.
      $$E(f) = \left[\begin{matrix} f_\L \\ f_\U \\ y_\L \\ \pi \end{matrix}\right]^T 
             \left[\begin{array}{c|c}
           \widetilde{D} + \lambda(D-A) & 
       -\widetilde{D} 
             \\\\ \hline\\
             -\widetilde{D} & 
             0 
             \end{array}\right] 
      \left[\begin{matrix} f_\L \\ f_\U \\ y_\L \\ \pi \end{matrix}\right] ,$$
      where $$\widetilde{D} = \left[\begin{matrix} D_\L & 0 \\ 0&\lambda w_0 D_\U\end{matrix}\right] .$$
      Let $Q = \left[\begin{matrix} I_\L & 0 \\ 0 & \lambda w_0 I_\U \end{matrix}\right]$. Then, $\widetilde{D} = DQ$.

    Notice that the energy function is convex in $f$, since the Hessian is diagonally dominant and symmetric.
      The gradient of the above expression w.r.t. $f$ is:
    $$\nabla_f E = (\widetilde{D} + \lambda(D-A))f - \widetilde{D}y' ,$$
      where $y' = \left[\begin{matrix} y_\L \\ \pi \end{matrix}\right]$. Setting this to 0, we get the global minimizer:
      \begin{tabbing}
      $f $\hspace{1mm}\=$= \inv{(\widetilde{D} + \lambda(D-A))}\widetilde{D}y'$\\\\
        \> $= \inv{(DQ + \lambda(D-A))}DQy'$\\\\
        \> $= \inv{(Q + \lambda(I-\inv{D}A))}Qy'$\\\\
        \> $= \inv{((Q + \lambda I)-\lambda\inv{D}A))}Qy'$\\\\
        \> $= \inv{(I-\lambda\inv{(Q + \lambda I)}\inv{D}A))}\inv{(Q + \lambda I)}Qy'$\\\\
    \> $= \inv{(I - B\inv D A)}(I-B)y' .$
      \end{tabbing}
      The last line can be readily verified given 
      $$B = \left[\begin{matrix} \frac{\lambda}{1+\lambda}I_\L & 0 \\ 0 & \frac{1}{1+w_0}I_\U \end{matrix}\right] .$$
    
    \subsection{Derivation of Initialization step}\label{sec:init_step}
      We have that $f = \inv{(I - R X^TX)}q$.
      The Kailath variant of the matrix inverse lemma gives:
      $$\inv{(A+BC)} = \inv{A} -  \inv{A}B\inv{(I+C\inv{A}B)}C\inv{A} .$$
      Using this, we have: 
      \begin{tabbing}
        $f $\=$= \inv{(I - R X^TX)}q$\\\\
        \> $=(I+(RX^T)\inv{(I-XRX^T)}X)q$\\\\
        \> $= q+(RX^T)\inv{(I-XRX^T)}Xq .$
      \end{tabbing}
    
    \subsection{Derivation of Updates}
      We have $\inv{K} = \inv{(I - XRX^T)}$ from last iteration.
      Further, only one element in $R$ changes: 
      
      $R^+ = R -\gamma e_i e_i^T$, 
      where $\gamma = -\left(\frac{\lambda}{1+\lambda}-\frac{1}{1+w0}\right)\inv{D_{ii}}$.
      Expanding $K^+$, we get:
      \begin{tabbing} 
        $K^+$ \= $:=I - XR^+X^T$\\\\
        \>$= K + \gamma X e_i e_i^T X^T$\\\\
        \>$= K + \gamma x_i x_i^T .$
      \end{tabbing}
      Here, we use Woodbury's Matrix inversion formula: 
      $$\inv{(A+UCV)} = \inv{A} - \inv{A}U\inv{(\inv{C}+V\inv{A}U)}V\inv{A} .$$
      From this, we have:
      \begin{tabbing}
    $\inv{(K^+)}$\= $= \inv{K} - \inv{K}(\gamma x_i)\inv{(1+ \gamma x_i^T\inv{K}x_i)}x_i^T\inv{K}$\\\\
      \>$=\inv{K} - \dfrac{\gamma\inv{K}x_i x_i^T\inv{K}}{1+\gamma x_i^T \inv{K} x_i}$\\\\
      \>$= \inv{K} - \dfrac{\gamma(\inv{K}x_i) (\inv{K}x_i)^T}{1+\gamma x_i^T \inv{K} x_i} .$
      \end{tabbing}
      Given that only one element in $q$ changes each iteration: $q^+_i = y_i \dfrac{1}{1+\lambda}$, our updated $f$ becomes
      $$f^+ = q^+ + R^+X^T\inv{(K^+)}Xq^+ .$$
      
      Each step in this process only involves $O(r^2 + rn)$ operations.
      Thus, this is the overall run-time per iteration.

	\subsection{Representation of $f$}\label{sec:ref_f}
        Here, we show how we can write $f$ as: \hspace{1mm}$f = \inv{M}Py'$
      \\\\where
      $P = \left[\begin{matrix}\dfrac{1}{\lambda} I_\L&0 \\ 0&w_0 I_\U\end{matrix}\right]D$, \hspace{3mm}$M = D + P - A$ 
      \\\\and $y'$ is, as before, the vector with true labels for labeled points and $\pi$ otherwise.
      From the derivation for the initialization, we can rearrange the terms and move some constants around as follows:
      
      \begin{tabbing}
      $f $\hspace{2mm}\= $= \inv{(DQ + \lambda(D-A))}DQy'$\\\\
        \> $= \dfrac{1}{\lambda}\inv{\left(D + \dfrac{1}{\lambda}DQ-A\right)}DQy'$ \\\\
        \> $= \inv{\left(D + \left(\dfrac{1}{\lambda}QD\right)-A\right)}\left(\dfrac{1}{\lambda}QD\right)y'$\\\\
    \> $= \inv{\left(D + P-A\right)}Py'$\\\\
        \> $= \inv{M}Py'$
      \end{tabbing}

   \subsection{Derivation of Impact factor}
   
   		The impact factor is defined to be the following:
        $$IM_i = f_i \sum\limits_{j \in \{U \backslash i\}} (f^+_j - f_j)$$
        where $f_j^+$ is the new score for $x_j$ after $x_i$ is labeled by the user.
        It is the change in $f$ over all unlabeled points, conditioned on the labeling $x_i$ as positive.
		Here is a brief description of this computation, followed by the detailed derivation.
        
        In order to compute this, we look at the change in $f$ given a positive label.

        We write $f = \inv{M}Py'$ as shown in \ref{sec:ref_f}.
        We can now reason about the change vector of $f$ given that unlabeled point $x_i$ is a positive: 
        $$\Delta f(i) := f^+ - f$$
        The following quantities can be computed directly from these change vectors.
        $$\Delta F = \left[\begin{matrix}\sum\limits_{j\in \U} \Delta f_j(1)\\ \vdots \\ \sum\limits_{j\in \U} \Delta f_j(n)\end{matrix}\right], \hspace{2mm}
        \Delta \widetilde{f} = \left[\begin{matrix}\Delta f_1(1)\\ \vdots \\ \Delta f_n(n)\end{matrix}\right]$$
        The $i^{th}$ element of the above vectors is computed assuming that $x_i$ has a positive label. 
        \textit{But}, we still take $x_i$ to be in $\U$ for the computation of $\Delta F$ because it simplifies computation, and can be corrected for.
        Further, only unlabeled entries matter in the computation, since the Impact Factor only depends on $\U$.

        The Impact Factor can be written as follows, where $\circ$ denotes the element-wise product:
        \begin{equation}
          IM = f \circ (\Delta F - \Delta \widetilde{f})
        \end{equation}
        We subtract $\Delta \widetilde{f}$ to correct for including $x_i$ in $\U$ in previous computations.
        We can shown that each term, and hence the whole Impact Factor, can be computed in $O(nr + r^2)$ time per iteration.
        The detailed steps follow below.\\\\

    From \ref{sec:ref_f}, we have that $f = \inv{M}Py'$.
    With this, we can write two equations:
      $$(D+P-A)f = P y'$$
      $$(D+P+\Delta P-A)f^+ = (P+\Delta P)y'^+ ,$$
    where $\Delta P = P^+ - P$.
	If we label point $x_i$ as positive, subtracting the second equation from the first, we get:
      \begin{tabbing}
        $(D+P-A)(f^+-f)$ \= $ = P(y'^+-y')+\Delta P (y'^+-f^+)$ \\\\
        \>$= \left[(y_i-\pi)P_{i,i} + \delta P (y_i-f^+_i)\right] e_i$
      \end{tabbing}
      with the $i^{th}$ standard basis vector $e_i$.
      The last equality is true for the following reasons. 
	  Since we labeled point $x_i$, we have $y'_i = \pi$ and $y'^+_i = y_i$ and for all $j \neq i$, $y'^+_j = y'_j$.
      Secondly, $\Delta P$ is 0 for every entry other than the $i^{th}$ diagonal entry, since only that element changes in $P$. 
      This tells us that when we label point $x_i$, we have:
      $$f^+ - f = \left[(y_i-\pi)P_{i,i} + \delta P (y_i-f^+_i)\right] \inv{M}e_i$$
      Substituting $y_i = 1$ for the Impact Factor and using the notation $\delta P = P_{i,i}^+- P_{i,i}$, we get:
      \begin{equation}\label{eq:dfi1}
        \Delta f(i) = [P_{i,i}^+ - \pi P_{i,i} - \delta P f_i^+]\inv{M}_{.,i}
      \end{equation}
      Using this, we can compute the two quantities of interest we defined before:
      $$\Delta F = \left[\begin{matrix}\sum\limits_{j\in \U} \Delta f_j(1)\\ \vdots \\ \sum\limits_{j\in \U} \Delta f_j(n)\end{matrix}\right], \hspace{2mm}
        \Delta \widetilde{f} = \left[\begin{matrix}\Delta f_1(1)\\ \vdots \\ \Delta f_n(n)\end{matrix}\right]$$
      Let $\Delta F_i := \sum\limits_{j\in \U} \Delta f_j(i)$ where the label of $x_i$ is taken as 1. 
      As mentioned before, $x_i$ is still included in $\U$.\\
      
      \noindent\textbf{Computing $\Delta \widetilde{f}$:} 
      In equation \ref{eq:dfi1}, we first solve for $f_i^+$ and subtract out $f_i$.
      This gives us the $i^{th}$ element of $\Delta f(i)$ as:
      \begin{equation}\label{eq:dfi2}
        \Delta f_i(i) = f^+_i - f_i = \dfrac{(P_{i,i}^+ - \pi P_{i,i} - \delta P f_i)\inv{M}_{i,i}}{1 + \delta P\inv{M}_{i,i}}
      \end{equation}

	  \noindent\textbf{Computing $\Delta F$:} 
      Notice that $\Delta F_i = \Delta f(i)^T u$ where $u$ is the indicator vector where $u_i = 1$ if $x_i$ is unlabeled and 0 otherwise.
      Using $(\inv{M}_{.,i})^T = \inv{M}_{i,.}$ since $\inv{M}$ is symmetric, equation \ref{eq:dfi1} gives us:
      \begin{equation}\label{eq:Dfi3}
	    \Delta F_i = [P_{i,i}^+ - \pi P_{i,i} - \delta P f_i^+]\inv{M}_{i,.}u
      \end{equation}
      
      Now, note that $P_{i,i}^+ = \dfrac{1}{\lambda}D_{i,i}$, since this denotes $P$ after $x_i$ is labeled.
      And, $P_{i,i} = w_0 D_{i,i}$ since $x_i$ is unlabeled at the start of this iteration.
      $P_{i,i}$ and $P_{i,i}^+$ do not depend on the label itself, just on whether $x_i$ is labeled or not. 
      With this, we can define new vectors which are the stacked versions of $P_{i,i}$ and $P_{i,i}^+$:
      $$\vec{L} := \left[\begin{matrix}P_{1,1}^+\\ \vdots \\ P_{n,n}^+ \end{matrix}\right] = 
      \dfrac{1}{\lambda}\left[\begin{matrix}D_{1,1}\\ \vdots \\ D_{n,n} \end{matrix}\right]
      \hspace{1mm} \text{ and } \hspace{1mm}
      \vec{U} := \left[\begin{matrix}P_{1,1} \\ \vdots \\P_{n,n} \end{matrix}\right] = 
      w_0 \left[\begin{matrix}D_{1,1}\\ \vdots \\ D_{n,n} \end{matrix}\right]$$
	  These vectors only are relevant in the indices of the unlabeled points. 
      Now, stacking equation \ref{eq:Dfi3} for all $i$, we get:
      \begin{equation}
        \Delta F = \left[\vec{L} - \pi\vec{U} - (\vec{L}-\vec{U})\circ(f + \Delta \widetilde{f})\right] \circ \inv{M} u
      \end{equation}
      
  	  If we are able to compute $\Delta \widetilde{f}$, $\inv{M}$ and $\inv{M}u$ efficiently, 
      then $\Delta F$ can be computed efficiently as it is just a constant number of point-wise $O(n)$ operations.
      Further, the Impact is also readily computed as: 
    \begin{equation}
      IM = f \circ (\Delta F - \Delta \widetilde{f})
      \end{equation}
      As noted before, we need to account for considering not removing $x_i$ from $\U$ while computing each component of $\Delta F$. 
      Thus, we have to subtract $\Delta \widetilde{f}$ before computing the final Impact Factor.
      
      In the following sections, we will discuss how to compute $\Delta \widetilde{f}$, $\inv{M}$ and $\inv{M}u$ efficiently.

      \subsubsection{Rewriting $\inv{M}$}
        We first show that $\inv{M} = \inv{(I - B\inv{D}A)}B\inv{D}$.
        Here, $B$ is the same as in section \ref{sec:minEf}.
		We know that $P=SD$ where 
        $$S = \left[\begin{matrix}\dfrac{1}{\lambda} I_\L&0 \\ 0&w_0 I_\U\end{matrix}\right] = \dfrac{1}{\lambda}Q$$
        Then, we have:
        \begin{tabbing}
          $\inv{M}$ \= $= \inv{(D + P - A)}$\\\\
          \>$=\inv{(D(I+S)-A)}$\\\\
          \>$=\inv{((I+S)-\inv{D}A)}\inv{D}$\\\\
          \>$=\inv{(I-\inv{(I+S)}\inv{D}A)}\inv{(I+S)}\inv{D}$\\\\
          \>$=\inv{(I-B\inv{D}A)}B\inv{D} .$ 
        \end{tabbing}
        The last step comes from the fact that $\inv{B} = I+S$.
        For our problem, $A = X^TX$. 
        Using the same steps as in Section \ref{sec:init_step}:
        \begin{equation}
            \inv{M}=\inv{(I-RX^TR)}R = (I+RX^T\inv{K}X)R
        \end{equation}
        where $R = B\inv{D}$ and $K = (I-XRX^T)$.
        The individual matrices in the above expression are already being computed in updates for $f$.

      \subsubsection{Computing $\Delta F$ given $\Delta \widetilde{f}$}
	  We have
      $$\Delta F = \left[\vec{L} - \pi\vec{U} - (\vec{L}-\vec{U})\circ(f + \Delta \widetilde{f})\right] \circ \inv{M} u$$
      Given $\Delta \widetilde{f}$, we can compute
      $\left[\vec{L} - \pi\vec{U} - (\vec{L}-\vec{U})\circ(f + \Delta \widetilde{f})\right]$
      in $O(n)$ time as it consists of only element-wise sums and multiplications.
      Further, $\inv{M}u$ can be computed as follows.
        \begin{itemize}
          \item $z = Ru$ changes only one element each iteration. 
          It can be initialized once and can be updated.
          \item $(I+RX^T\inv{K}X)z$ can then be computed in $O(rn)$ time by cascading the matrix-vector multiplication in.
        \end{itemize}
    These operations take $O(rn)$ time per iteration.

    \subsubsection{Computing $\Delta \widetilde{f}$ via updates}
		We have
        $$\Delta \widetilde{f} = \left[\begin{matrix}\Delta f_1(1)\\ \vdots \\ \Delta f_{n}(n)\end{matrix}\right]$$
        where each element is given by equation \ref{eq:dfi2}. 
        This can be written as:
        \begin{equation}
          \Delta \widetilde{f} = \left[\vec{L} - \pi\vec{U} - (\vec{L}-\vec{U})\circ f\right] \circ diag(\inv{M}) \circ \inv{N}
        \end{equation}
        where $N = diag\left(\mathbb{1}+(\vec{L}-\vec{U})\circ diag(\inv{M})\right)$.
    	We compute $diag(\inv{M})$ as follows:
        \begin{tabbing}
          $diag(\inv{M})$ \= $=diag ((I+RX^T\inv{K}X)R)$\\\\
          \>$=diag (I+RX^T\inv{K}X)\circ diag(R)$\\\\
          \>$=(\mathbb{1}+diag (RX^T\inv{K}X))\circ diag(R)$
        \end{tabbing}
          Thus,
        \begin{equation}
          diag(\inv{M}) = \left(\mathbb{1}+diag(R) \circ diag (X^T\inv{K}X)\right)\circ diag(R)
        \end{equation}
    	From this, we see that we just need to store and update $J = diag (X^T\inv{K}X)$ every iteration.
    	Here's how we can do this:
        \begin{itemize}
          \item We initialize $j_i= x_i^T \inv{K} x_i$ and 
            $J = \left[\begin{matrix}j_1\\ \vdots \\ j_n\end{matrix}\right]$ at the start. 
            This computation takes $O(nr^2)$ time once we have $K$.
          \item Then, as we update $K$, we also update $J$. 
            Here, $i$ is the index of the point to be labeled and $t$ is the index of $J$ being modified.
            \begin{tabbing}
              $j_t^+$\= $= x_t^T \inv{(K^+)} x_t$\\\\
              \>$= x_t^T\left(\inv{K} - \dfrac{\gamma(\inv{K}x_i) (\inv{K}x_i)^T}{1+\gamma x_i^T \inv{K} x_i}\right)x_t$
            \end{tabbing}
          Therefore,
            $$j_t^+ = j_t - c\cdot (x_t^T(\inv{K}x_i))^2$$
          where $c = \dfrac{\gamma}{1+\gamma x_i^T \inv{K} x_i}$ is already computed in the updates to $f$. 
          Since $(\inv{K}x_i)$ is also computed while updating $f$, updating each element $j_t$ only takes $O(r)$ time because it consists of only a dot product and a constant number of scalar operations.
        \end{itemize}
      Thus, updating the entire $J$ vector only takes $O(nr)$ per iteration.
      Once we have $J$, we can compute $diag (\inv{M})$ in $O(n)$ time through element-wise operations.
	  
      Finally, after computing $diag(\inv{M})$, we can compute $\Delta \widetilde{f}$ in $O(n)$ time every iteration, 
      again using only element-wise sums and products.
        
      \subsubsection{Putting it all together}
    	We showed that in computing both $\Delta F$ and $\Delta \widetilde{f}$, we need only $O(nr^2)$ initialization time and $O(nr + r^2)$ time per iteration of the algorithm.
    	Then, putting them together to get the Impact Factor takes only $O(n)$ time.
        This is the same as what is needed to compute the initialization and updates for $f$.
        
        In this way, the Impact Factor computation also scales with the other computations performed in the algorithm.
    
    \section{Proofs}
    \subsection{Proof of Lemma 4.1}
      \noindent The adjacency matrix $A$ is assumed to only have positive values.
      
      \noindent\textbf{Lemma 4.1}
      
      \noindent {\it Let $A = A_1 + A_2$ where $A_1$ is the block diagonal component of the similarity matrix and $A_2$ is the pure cross-similarity component.

       If $||A_2||_1 < \epsilon$, then $||\widetilde{M}_{12}||_1 < \left(\dfrac{1}{c\cdot d_{min}}\right)^2\epsilon$ 
        where $c=\min\{\frac{1}{\lambda}, w_0 \}$ and $d_{min}$ is the minimum degree in the graph.}\\\\
        
      \noindent\textbf{Proof:}
      We want to show that if $A$ is ``close'' to block diagonal, then $\inv{M}$ is close to block diagonal. 
      We have that:
      \begin{tabbing}
       $M$ \= $=D+P-A$\\
       \>$=D+P-A_1-A_2$\\
       \>$=M_1-A_2$
     \end{tabbing}
      where $M_1=D+P-A_1$ is block diagonal.
      Using Woodbury's Matrix Inversion lemma, we have:
      \begin{tabbing}
       $\inv{M}$ \= $=\inv{M_1} + \inv{M_1}A_2\inv{(I-\inv{M_1}A_2)}\inv{M_1}$\\\\
       \> $=\inv{M_1} + \inv{M_1}A_2\inv{(M_1(I-\inv{M_1}A_2))}$\\\\
       \> $=\inv{M_1} + \inv{M_1}A_2\inv{(M_1-A_2)}$
     \end{tabbing}
    Thus, $\inv{M}-\inv{M_1} =\inv{M_1}A_2\inv{(M_1-A_2)}$ where we have that $\inv{M_1}$ is block diagonal.
      Taking the 1-norm of this, we have from sub-multiplicity of induced matrix norms:
      \begin{tabbing}
        $||\inv{M}-\inv{M_1}||_1$ \=$= ||\inv{M_1}A_2\inv{(M_1-A_2)}||_1$\\\\
        \>$\le ||\inv{M_1}||_1\cdot||A_2||_1\cdot||\inv{(M_1-A_2)}||_1$
      \end{tabbing}
      Here, we use a result by Varah \cite{varah} which is as follows:\\\\
      \noindent For a diagonal dominant $n \times n$ matrix $J$, 
      $$||\inv{J}||_{\infty} \leq \max\limits_{1\leq i \leq n} \frac{1}{\Delta_i(J)}$$
      where $\Delta_i(J)$, is the $i^{\rm th}$ diagonal dominance defined by
      $$\Delta_i(J) := |J_{ii}| - \sum\limits_{j \neq i}|J_{ij}|,~i=1,\hdots,n.$$
      For $M = D+P-A$, we have $\Delta_i(M) = P_{ii}$ as the contribution of $D-A$ to $\Delta_i(M)$ is 0.
      The minimum value of $\Delta_i(M)$ is then lower bounded by $c\cdot d_{min}$.
      Further, since $A_2$ only introduces off diagonal elements to $M = M_1-A_2$, we have that 
      $\Delta_i(M_1) \geq \Delta_i(M)$ for all $i$.
      Thus, the minimum value for $\Delta_i(M_1)$ is also lower bounded by $c\cdot d_{min}$.
      This means that:
      $$||\inv{M}||_\infty \leq \dfrac{1}{c\cdot d_{min}} \textnormal{ and }
      ||\inv{M_1}||_\infty \leq \dfrac{1}{c\cdot d_{min}}$$
      Since $\inv{M}$ and $\inv{M_1}$ are symmetric, we have that 
      $$||\inv{M}||_1=||\inv{M}||_\infty \textnormal{ and }
      ||\inv{M_1}||_1=||\inv{M_1}||_\infty$$
      as the 1-norm is the maximum absolute row-sum and the $\infty$-norm is the maximum absolute column sum.
      
      This finally gives us:
       \begin{tabbing}
        $||\inv{M}-\inv{M_1}||_1$ \=$\le ||\inv{M_1}||_1\cdot||A_2||_1\cdot||\inv{(M_1-A_2)}||_1$\\\\
        \>$< \left(\dfrac{1}{c\cdot d_{min}}\right)^2 \epsilon$
      \end{tabbing}
    	
        If we write $\inv{M} = N_1 + N_2$ where $N_1$ is the block diagonal component of $\inv{M}$ and $N_2$ is the off block diagonal component, we have that:
    $$||(N_1 -\inv{M_1}) + N_2||_1 < \left(\dfrac{1}{c\cdot d_{min}}\right)^2 \epsilon$$
      Further, we have that $||N_2||_1 < ||(N_1 -\inv{M_1}) + N_2||_1$ since we are just introducing numbers where $N_2$ is 0. 
      This can only increase the absolute row-sums.
      With this, we have that $$||N_2||_1 < \left(\dfrac{1}{c\cdot d_{min}}\right)^2 \epsilon $$
      Since $N_2$ is the block diagonal component of $\inv{M}$, it is exactly the following as defined in Section \ref{sec:goodsim}:
      $$N_2 = \left[\begin{matrix}0&\widetilde{M}_{12} \\\widetilde{M}_{12}^T &0 \end{matrix} \right]$$
      This gives us that $$||\widetilde{M}_{12}||_1 \leq ||N_2||_1 < \left(\dfrac{1}{c\cdot d_{min}}\right)^2\epsilon$$
      $\hfill\blacksquare$